# Anomaly Detection Support Using Process Classification

Sebastian Eresheim,[*] Lukas Daniel Klausner,[†] Patrick Kochberger[†]
sebastian.eresheim@fhstp.ac.at, mail@l17r.eu, patrick.kochberger@fhstp.ac.at
[*] *Josef Ressel Center for Unified Threat Intelligence on Targeted Attacks*
[†] *Institute of IT Security Research*
*St. Pölten University of Applied Sciences*

*Abstract*—Anomaly detection systems need to consider a lot of information when scanning for anomalies. One example is the context of the process in which an anomaly might occur, because anomalies for one process might not be anomalies for a different one. Therefore data – such as system events – need to be assigned to the program they originate from. This paper investigates whether it is possible to infer from a list of system events the program whose behavior caused the occurrence of these system events. To that end, we model transition probabilities between non-equivalent events and apply the $k$-nearest neighbors algorithm. This system is evaluated on non-malicious, real-world data using four different evaluation scores. Our results suggest that the approach proposed in this paper is capable of correctly inferring program names from system events.

## 1. Introduction

Targeted attacks are no longer just on the horizon – they have become a real threat to companies and even nation states. According to Symantec [1], the number of organizations affected by targeted attacks increased by 10% in 2017 and over 140 known targeted attack groups are currently active, many of them suspected of being nation state sponsored. One of the main motives of over 90% of these groups is intelligence gathering. This means once attackers have gained access to a network, their intent is to collect as much information about their victim as possible. Therefore they need to spread across the network and try to stay on it as long as possible. In order to do this, stealth techniques are required to avoid the victim from noticing the attack.

While spawning an unknown process on a victim system for remote interaction is a rather conspicuous task, using already running processes or starting a well-known non-malicious process and executing commands in its context is a more sneaky way to do so. This is also the reason why two artificially crafted programs with different names, but almost indistinguishable behavior are not a use case of interest for us, because programs which do not naturally occur on a victim's computer are already suspicious enough by themselves. But by misusing an already running process for malicious purposes, the malicious activity might remain undetected by the user. This abuse of running processes will, however, cause deviations in the behavior of that particular process. An analyst might be able to reveal these behavior dissimilarities in a dynamic executable analysis. The vast amount of data spawned by today's systems requires automation to properly accomplish this task. Machine learning (ML) algorithms provide tools to cope with large amounts of unknown, but similar to already known, data and enable this kind of decision support.

This paper investigates whether it is possible to infer from a list of system events the program whose behavior caused the occurrence of these system events. This is of interest for two reasons:

First, in a supervised context, program behavior anomalies can be detected when the behavior of a particular process is classified to be from a different program than the one it actually was created by; and second, in an unsupervised context, given the execution of an unknown program, similarities to known programs provide additional information to the analyst. To that end, we introduce a classification system mapping chronological lists of system events to the programs which are most likely to have a behavior that generates these event lists and then evaluate this system via unseen data.

Assigning the program name to system events is a vital step in the automation of attack detection. On the one hand, it is a preliminary stage to an anomaly detection system where behavior anomalies for each program are detected; on the other hand, its outcome can already be an indication of a compromise. For example, if an attack alters the event sequence of process A so that it appears to be an event sequence of process B instead, then the proposed classification system is capable of identifying the mismatch between natural and classifier-identified process name. This is possible because the executable file name of process A suggests the system events come from an execution of process A, while the classification system identifies the origin of the system events to be an execution of process B. Therefore, this mismatch between natural and classifier-identified process name can already indicate that the integrity of a process has been compromised.

In order to answer the main question of this paper, we first use event transition probabilities from Markov chains to model the event-based environment and later apply an ML algorithm to classify the programs accordingly. The class a system event list is assigned to corresponds to the name of the program which is considered to be the most likely origin of the system event list.

The rest of this paper is structured as follows: section 2 elaborates on related work, section 3 introduces the necessary background knowledge and the techniques used in the proposed classifier, section 4 gives an overview of the proposed system, section 5 evaluates the system, section 6 discusses the outcome and section 7 concludes this paper.

## 2. Related Work

The goal of the proposed system is to assign a program name to a list of generated events of a specific process. The broader objective, however, is to check whether a process has been compromised, shows malicious behavior and therefore deviates from the usual behavior of the program; in other words, whether the integrity of the process behavior can be guaranteed or not.

Early process behavior integrity checks were limited to integrity checks of the code of an executable and include tools such as Tripwire [2], which computes cryptographic hash values of code segments or a whole file and compares them to a secure baseline. A more modern approach to ensure integrity is code signing, wherein an executable is signed with a digital certificate. This way the operating system (OS) ensures the executable's code has not been altered and also provides information about the code author. Both approaches check the code of a process before it is first executed. This is unsatisfying in scenarios where already running applications are exploited and therefore code might change during execution.

This integrity-check problem belongs to a problem class called *time-of-check, time-of-use* (TOCTOU) where an attacker modifies a variable or an object after it has been checked for certain validity conditions but before it is used in the execution. Bratus *et al.* [3] give an example for a TOCTOU vulnerability in trusted computing.

In the context of digital rights management (DRM) the concept of tamper-proof software [4] is frequently used (referring to guaranteeing that a program executes as intended). One method of software tamper-proofing is using integrity checks of code that is already loaded into a running process. The process uses parts of its own code as input for a hash function and compares the output to previously computed values of untampered code. Even though the code is checked within the context of the process, during execution, the TOCTOU problem is not evaded, because this approach still focuses on the code loaded into the process and not on the code that is actually computed. For example, Hund *et al.* [5] succeeded in creating return-oriented malware that alters process behavior without changing the code of a process. Thus such integrity checks are also not sufficient to ensure that processes execute as intended.

Another research area that has a common ground with the problem of process behavior integrity is malware detection (cf. [6] [7], [8], [9]). However, the main focus in malware detection is to determine whether the primary intent of an executable is malicious or not, generally under the assumption that the program's behavior does not change. It is mostly a classification task on whether an executable is capable of bad/malicious behavior or not depending on certain standards. In contrast, the problem of process behavior integrity deals with mostly non-malicious behavior that dynamically becomes malicious only when the process is attacked.

In the field of anomaly detection for intrusion detection, building on the seminal work of Forrest *et al.* [10], many solutions have already been introduced to tackle the intrusion problem. In contrast to many of these known solutions (e.g. [11], [12], [13], [14]), we do not use system calls as the basis of the automated analysis; we use system events instead, which occur on a higher level and therefore contain less specific information when compared to system calls. The second distinction from many existing works in this field is that we do not only try to detect anomalies from expected behavior and data, but rather try to identify a program based on its generated events, which solves a more general and more complicated task.

As mentioned above, anomalies in intrusion detection depend on the process they occur in. According to the taxonomy of Chandola *et al.* [15], these anomalies hence belong to the class of *contextual anomalies*, and consequently, two kinds of attributes are required for detection of such anomalies: *contextual attributes* (to determine the context) and *behavioral attributes* (to identify the anomalies themselves). The purpose of the classification system proposed in this paper is to provide the contextual attribute (namely the program) from system events.

Markov chains (subsection 3.1) respectively transition probabilities have already been proved to be useful features in previous work. Farag [16] used Markov chains to model a sequence of directional strokes for cursive script recognition. He achieved his goal by feeding the probabilities of the transition matrix to a maximum likelihood classifier. Hassanpour *et al.* [17] modeled the texture of images of banknotes using Markov chains for paper currency recognition. Ahmed *et al.* [18] modeled temporal information of API call sequences with time-discrete Markov chains. They determined the most relevant elements of the transition matrix of the Markov chain via the measure of information gain and utilised these elements as features for ML algorithms. Their setup results in a runtime malware analysis and detection scheme that is based solely on *memory management* and *file I/O* API calls. Wang *et al.* [19] also built their feature extraction system for image tampering detection employing Markov chains. Based on the transition probability matrix of a thresholded edge image, they calculate a stationary distribution which is transformed into a feature vector for a support vector machine. Rafique and Abulaish's xMiner [20] utilises transition probability matrices of multi-order Markov chains as a tool for feature extraction on byte-level network traffic. Combined with a principal component analysis and five different supervised ML algorithms (one of them *k*-nearest neighbors), it is capable of detecting vulnerability exploits in network traffic. Rafique *et al.* [21] extracted transition matrices of Markov chains as features for SMS spam detection. Paired with four different evolutionary algorithms as well as four supervised ML algorithms, they show that transition probabilities are a feasible feature for SMS spam detection. García [22] calculates the absolute differences of elements of transition probability matrices of Markov chains. The author compares network connections of possible botnet clients to a certain threshold to classify traffic as botnet-related or benign. Rafique *et al.* [23] created state transition matrices similar to transition probability matrices of Markov chains to classify malicious network traffic. Additionally, they applied the gain ratio feature selection scheme to chose the most discriminative transitions in the matrix. Previous work of Marschalek *et al.* [24] included transition probability matrices from Markov chains for distance calculation. Instead of utilizing it for program classification, they apply it in combination with distance-based clustering for separating malware from benign software and evaluated it with a set of mostly benign data.

# 3. Preliminary

This chapter summarises concepts and techniques and briefly describes terms which are used in the following chapters.

## 3.1. Markov Chains

A (time-discrete) Markov chain [25] is a sequence of random variables such that the outcome of one random variable is only affected by its immediate predecessor. Formally, this property is defined as

$$P(\xi_{n+1} = j \mid \xi_0, \ldots, \xi_n) = P(\xi_{n+1} = j \mid \xi_n) \quad (1)$$

and is also known as the *Markov property*. In this paper, the set $S = \{c_1, \ldots, c_N\}$ of possible values for the random variables, the so-called *state space*, is always finite. The one-step transition probability of a Markov chain from state $i$ to state $j$ is

$$p_{ij} = P(\xi_{n+1} = j \mid \xi_n = i).$$

The full transition matrix $P$ for a Markov chain with the one-step transition probabilities $p_{ij}$ is the $N \times N$ matrix

$$P = \begin{pmatrix} p_{11} & p_{12} & \cdots & p_{1N} \\ p_{21} & p_{22} & \cdots & p_{2N} \\ \vdots & \vdots & \ddots & \vdots \\ p_{N1} & p_{N2} & \cdots & p_{NN} \end{pmatrix}.$$

The matrix $P$ satisfies the conditions

$$0 \leq p_{ij} \leq 1 \quad \forall\, i,j \in \{1, \ldots, N\} \quad (2)$$

and

$$\sum_{j=1}^{N} p_{ij} = 1 \quad \forall\, i \in \{1, \ldots, N\}. \quad (3)$$

Furthermore, any matrix that satisfies these two conditions is called a *Markov matrix* and can be viewed as the transition matrix of a Markov chain.

## 3.2. Principal Component Analysis

The method of *principal component analysis* (PCA) [26, 27] is used in architectures with a high-dimensional feature space. High-dimensional feature spaces are often problematic because they require either a lot of memory and/or processing power in order to be processed or contain a lot of noise within the data which may negatively affect the classification results. Therefore, it is necessary to separate the relevant data from noise in order to efficiently compute the classification. PCA achieves this by projecting the data onto a linear subspace of the original space.

For this method, it is assumed the data is *centered*, which means the mean value of the data should be zero. (If this is not already the case, it can easily be achieved by subtracting the mean value from each data point.) In a first step, the covariance matrix of all features is computed. For a system with $n$ features $X_1, \ldots, X_n$, this yields the matrix

$$C := (c_{i,j})_{i,j \in \{1,\ldots,n\}}$$

where $c_{i,j} := \mathrm{cov}(X_i, X_j)$ and where $\mathrm{cov}(X_i, X_j)$ denotes the covariance of $X_i$ and $X_j$. Calculating the eigenvectors and eigenvalues of this matrix, sorting them according to the eigenvalues' absolute values and then selecting only the largest $m$ ones yields a $m \times m$ matrix and the associated $m$-dimensional linear subspace. By this calculation, the eigenvalues of the covariance matrix are exactly the variances of the features, which is precisely why those linear combinations of features with the most variance have been selected. Because the eigenvectors are by definition orthogonal, the $m \times m$ matrix is also orthogonal, which is why the eigenvectors can be seen as a rotation of (some of) the axes. Projecting all data points onto this lower-dimensional subspace yields a data set with most of the information, less of the noise and a reduced number of dimensions.

## 3.3. $k$-Nearest Neighbors

The *k-nearest neighbors* ($k$-NN) classifier [28] is one of the oldest and best known ML algorithms. It essentially operates in two steps:

1) Find the $k$ nearest neighbors to a new data point.
2) Determine the classes of those neighbors, determine the most common class by some majority voting procedure and assign the new data point to this class.

The set of nearest neighbors depends on the distance function that is used to calculate the distance between data points. In many cases the Euclidean distance is used for this purpose, although other metrics can also be considered. The distance might also affect the majority voting in the second step, because it can be held in two different ways. The simpler procedure weights every neighbor's vote equally and picks the most commonly occurring class. Alternately, each neighbor's vote can be weighted by its distance to the new data point, meaning closer neighbors have more impact on the classification than more distant ones. (Ideally, the value $k$ which determines the number of neighbors to be considered is not a multiple of the number of classes, to avoid tied votes and thus the need for tiebreaks.)

## 4. Proposed Architecture

This section elaborates on the proposed system: We first provide a brief overview in subsection 4.1 and then specify the details of the different steps of the process: data transformation in subsection 4.2, feature extraction in subsection 4.3, dimensionality reduction in subsection 4.4 and finally classification in subsection 4.5.

### 4.1. Overview

The underlying data for the proposed system are system events which occur when a process interacts with the OS. These events are higher-level information, which is already a first abstraction from the underlying system calls. In our data set, a single event consists of a type, a subtype, a value and a timestamp. The event type is restricted to one of five possible values:

- process event
- file event
- image load event
- registry event
- network event

The subtype is a numerical value dependent on the event type and holds more specific information about the process–OS interaction. For example, it encodes whether the operation is a *read* or *write* operation. The event value also depends on the event type; for the first four event types, it is a path to either a file or to a registry key, while for network events, the value is an integer (the size of the involved data in bytes).

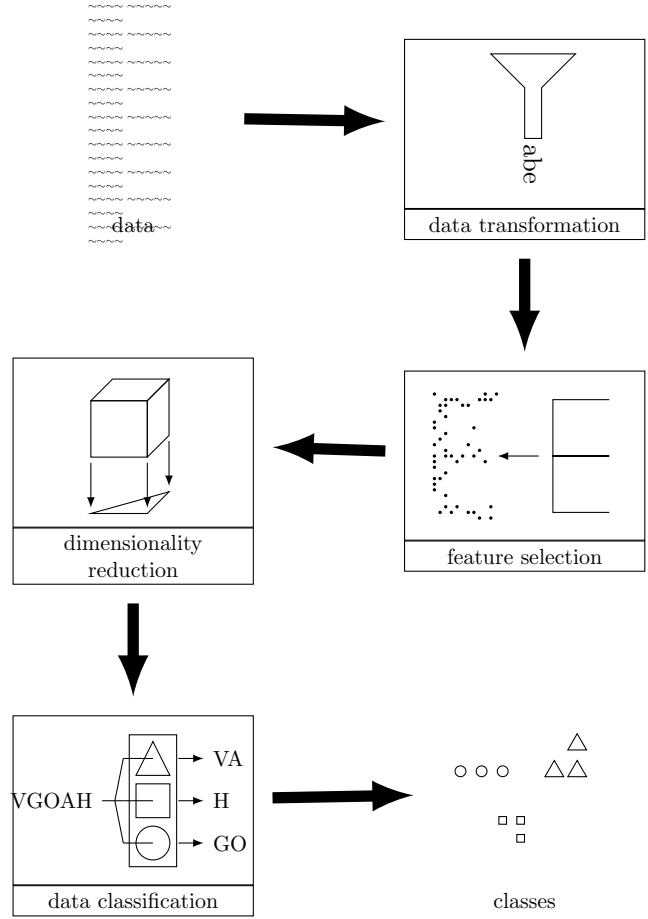

Figure 1. Overview of the proposed system. In the first step, the data (given as one list per process) is reduced to one string per process. In the second step, the string is transformed into numerical, high-dimensional data; hence dimensionality reduction is required as a third step before classification is feasible in the final step.

The proposed system consists of four steps, which are displayed in Figure 1.

The first step is a *data transformation* step (subsection 4.2) to better cope with the vast amount of data. We transform the data by introducing equivalence classes of events. For ease of use, we represent each equivalence class by a character, thus transforming a list of events into a string of characters (i. e. event equivalence classes). The motivation for this is that a human observer might already be able to see (non-)similarities between process executions by visual inspection of these strings. The *feature extraction* step (subsection 4.3) interprets the strings as outcomes of Markov chains and transforms them into their corresponding tran-sition probability matrix. The elements of the transition probability matrix are then used as features for the ML algorithm. Because using the transition probability matrices as feature vectors creates a very high-dimensional feature space, we apply *dimensionality reduction* (subsection 4.4) to improve the classification quality as well as reduce the required computing resources. In the final *classification* step, we apply the supervised $k$-NN classifier (subsection 4.5) to the data.

### 4.2. Data Transformation

The goal of the first step is to transform the data into a more suitable, but still human-readable format. Specifically, a human observer should be able to decide by visual inspection whether some process executions are generally similar or dissimilar, without necessarily being able to interpret every single character of the string. We achieve this by building *equivalence classes* of events. Two events are considered equivalent when both meet a certain condition. These conditions depend, among other things, on the event type; for example, two file events are equivalent if both of their paths start with "`C:\Windows`". The data transformation also helps to prevent overfitting by further generalising the underlying information. For ease of use, each equivalence class is assigned a character. Since every equivalence class has its own character, we can consider the family of equivalence classes as an *alphabet A*. Substituting each event with its representative character from the alphabet transforms an event list into a string, thus reduces the amount of data needed for storage and makes it a more easily comprehensible model of process–OS interactions. Conversely, every string resulting from this transformation is an abstraction of a chronological series of system events.

The equivalence relations underlying the aggregation into equivalence classes depend on the type of the system event. Table 1 displays the distinction criteria corresponding to each event type as well as the letter ranges (or corresponding UTF-8 hex values) that are used in the alphabet. The criteria are abbreviated as follows: "path" is either the file or registry key path, "subtype" is the event subtype as stated in subsection 4.1, "size" refers to the data size in bytes, "home directory" describes whether a file resides in the same directory as the executable of the process generating the event (or a subdirectory thereof) or not. For image load events, we were able to assign DLL files included in Windows to specific Windows functionalities like file system, networking, I/O, etc. and also used these as a distinction criterion for image load events (subsumed under the term "Windows functionality").

| event type | character ranges | distinction criteria |
|---|---|---|
| process | A–D, a–d | path |
| registry | 170–183 | path, subtype |
| image load | J–L, j–l, C0–16D | path, home directory, Windows functionality |
| file | 184–1CB | path, subtype, home directory |
| network | R, r, u–x | size, subtype |

Table 1. Overview of the alphabet, including which features were chosen for building equivalence classes and which letters are used.

Besides the system events, time is also included as a crucial factor in our model. Time is semantically important to discern between processes where system events happen immediately after each other and processes which sleep or wait for certain conditions in between events. Therefore we also introduce *time characters*, which are a set of characters whose purpose is to denote time intervals without system events taking place. Table 2 shows the *time characters* and the process idle intervals they represent.

| character | idle time elapsed |
|---|---|
| . | 1 millisecond |
| , | 10 milliseconds |
| + | 100 milliseconds |
| : | 1 second |
| ^ | 10 seconds |
| - | 1 minute |
| _ | 10 minutes |
| # | 1 hour |
| ~ | 1 day |

Table 2. Time characters and corresponding process idle time elapsed.

This approach is more practical than repeating a single time character, since e.g. having only a "1 millisecond" time character would require 1000

```
regedit.exe    C.....ÐĤ....ĦÐÐ.ÈIJĆîŢÐкІІ·κ..ÐÂŖIJĬ.ÀŪIJĤ.ƖØ.ŢŌú,Ō...К
regedit.exe    C.....ĤÐ..ÐÐIƉŢÐк.І‍·кÈ.IJĆ.î.ÂŖIJĬ.Ð.ĤÀŪ.IJ.Ø.Ɩ.Ō.Ţú,,,,Ō
regedit.exe    C.ÐĤ.ÐÐ.ÐĤŢĤкÐĆÔĦÐIJĬ.ІĦØкÔØöŤŦ.ĊÈîŖ.ÂIJ.ĤĆÀŪ.Ɩ,Ţ
regedit.exe    C^^^^K,,,K__ð,Ĥ.Ĥ-K,K::::K,K---žžžž
WerFault.exe   A,ĤÐÐ.ÐIŢÐĆĤкŢæŁ.æIJĬÎƉÐúIJ..Ĥ
WerFault.exe   A,ĤÐÐÐ.IŢÐĆĤкŢæŁæ.IJĬÎƉÐúIJ..Ĥ
WerFault.exe   A,ĤÐÐÐ.IŢÐĆĤкŢæŁæIJĬÎ.ÐúIJ..Ĥ
WerFault.exe   A........ĤÐ.ÐÐIŢÐĆĤкŢæ.ŁæIJĬÎƉÐúIJ..Ĥ
WerFault.exe   A......ĤÐÐÐ..IŢÐĆĤкŢæŁæIJĬÎƉÐú.IJ..Ĥ
WerFault.exe   A....ĤÐ.ÐÐIŢÐĆĤкŢæ.ŁæIJĬÎƉÐúIJ..Ĥ
WerFault.exe   A...ĤÐÐÐ.IŢÐĆĤкŢæŁæIJĬÎƉÐú.IJ
WerFault.exe   A..ĤÐ.ÐÐ.IŢÐĆĤкŢæŁ.æIJĬÎ.Ðú..IJ
WerFault.exe   A..ĤÐ.Ð.Ð.IŢÐĆĤкŢ..æŁ..æIJĬÎƉÐú.IJ
WerFault.exe   A..ĤÐ.ÐÐIŢÐĆĤ.кŢæŁ..æIJĬÎ.Ðú.IJ
WerFault.exe   A..ĤÐÐÐ.IŢÐĆĤкŢæŁ...æIJ.ÏƉÐúIJ...Ĥ
WerFault.exe   A..Ĥ.ÐÐÐ...IŢÐ.ĆĤкŢæŁæIJĬÎƉÐú.IJ
WerFault.exe   A.ĤÐ.ÐÐ..IŢÐĆĤкŢæŁæ..IJĬÎƉÐúIJ...Ĥ
WerFault.exe   A.ĤÐÐ.ÐIŢÐĆĤкŢæŁ.æIJĬÎƉÐúIJ..Ĥ
WerFault.exe   A.ĤÐÐÐ.IŢÐĆĤкŢæŁ..æIJĬÎƉÐúIJ..Ĥ
xcopy.exe      B.ÐđĥĥđÐđĤ.ţĥÐÐ.ÐÉî.Ŧ..ÏIJÎIJÔÔÐÔÐ.Ð.ÐŤĤþØƖ.ŦкŢƉŢ.Qđ
xcopy.exe      B.ÐđĥĥđĤÐ.đţĥ.ÐÐ.ÐÉî.Ŧ..ÏIJÎIJÔÔÐÐÐ.ÐŤþĤƖØ.ŦкŢƉŢ.Qđ
xcopy.exe      B.ÐĥĥđđĤ.ÐđţĥÐÐ.ÐÉî.Ŧ.Ð.ÏIJÎIJÔÔÐÐÐ.ÐŤþĤƖØ.ŦкŢƉŢ..Qđ
xcopy.exe      B.ĤđÐ.....đÐĥĥđ.ÐţĥÐ.ÉÐî.Ŧ.IJĬ.ÔÐƖÐØIJÔ.ĤŤþŦк.NŢŢ
xcopy.exe      B.ĤđÐ....ĥĥ.đÐđţĥ.ÐÐ.ÐÎÉ.Ŧ..IJĬÏIJÔÔÐ.ƖØÐŤþĤ.ŦкNŢŢ
xcopy.exe      B.ĤđÐĥ.ĥđÐđţĥÐÐ.ÐÉî.Ŧ..IJĬÏIJÔÔÐ.ƖØÐŤþĤ.ŦкNŢŢ
xcopy.exe      B.ĤđÐĥĥ.đÐđţĥÐÐ..Ð.ÐÉî.Ŧ..IJĬÏIJ.ÔÔÐƖØ.ÐŤĤþŦ.кNŢ.Ţ
```

Figure 2. Example strings for executions of the programs `regedit.exe`, `WerFault.exe` and `xcopy.exe`. Strings like these are the outcome of the data transformation step, yielding one string per process.

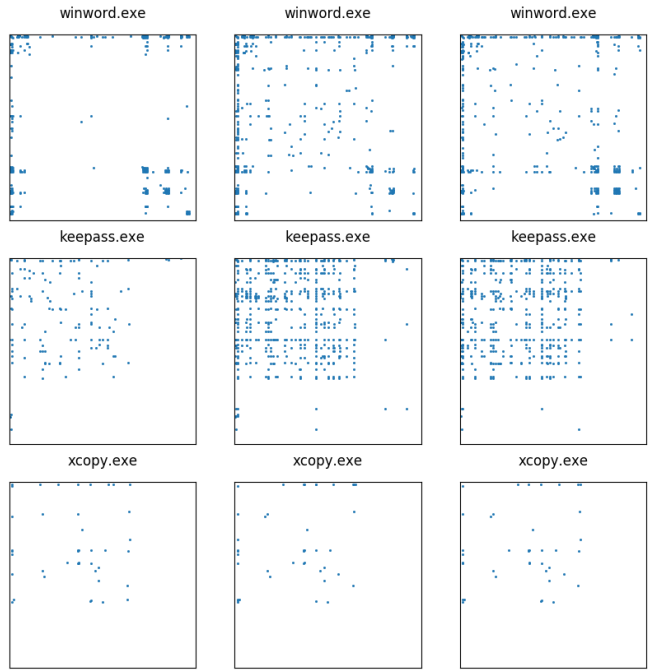

Figure 3. Transition matrices of strings from three programs show that matrices from the same program look more similar than matrices of different programs. Each of the nine images corresponds to a transition matrix, with each pixel corresponding to a matrix element; white pixels correspond to zero-valued elements, while blue pixels correspond to non-zero-valued elements.

successive characters of this time to denote an idle time of 1 second, whereas our approach requires just one character. Since processes might occasionally idle for days or even weeks, the string length would become unmanageable without a hierarchy of time characters.

Figure 2 shows example strings for executions of three different programs, namely `regedit.exe`, `WerFault.exe` and `xcopy.exe`. Even though these examples are particularly short-lived processes, it is easy to see that different executions of the same program are quite similar, whereas executions of distinct programs look dissimilar.

### 4.3. Feature Extraction

The first step (subsection 4.2) introduced an event-list-to-string transformation resulting in one string per process execution. Because classifiers usually operate on numerical values, we require a second transformation. We treat each string as the outcome of a sequence of random variables. Since the outcome is already known, we calculate the one-step probabilities for all characters appearing in the string. The resulting matrix fulfills the conditions in Equation 2 and Equation 3 and can therefore be interpreted as a probability matrix of a Markov chain. After applying this procedure to multiple strings from the previous step, we observed that transition matrices from strings of the same program look more similar compared to transition matrices from strings of other programs, even though the strings themselves do not always actually fulfill the Markov property (Equation 1). Figure 3 shows an excerpt of these observations in the form of transition matrices from three different programs. Each image in the figure corresponds to a transition matrix, and each pixel therein corresponds to one matrix element. Colored pixels denote that the matrix has a non-zero value at the corresponding position, while white pixels denote that the position's value in the matrix is zero. We can now apply ML algorithms to these matrices with the matrix elements as the feature vector elements. Because they are quite large in size, the next step is vital to achieve good classification results in a reasonable

## 4.4. Dimensionality Reduction

It is well known that $k$-NN calculation is imprecise in high-dimensional spaces (cf. Beyer et al. [29]), which is why we apply dimensionality reduction to the extracted features. Due to the sparsity of the transition probability matrix, a standard PCA is not feasible for two reasons. First, PCA computes the full covariance matrix of the feature vector, which has size $N^2$ for feature vectors of length $N$, and the feature vector length $N$ itself is already the square of the alphabet size $|A|$. Second, PCA requires centered data, forcing addition/subtraction operations on many elements of the feature vector. Since most of them are zero initially, a sparse format is suitable to store them in memory and thereby save a lot of storage space; but adding/subtracting operations on many elements of the feature vector turns the initial sparse vector into a dense vector and adds a lot of costly storage requirements.

Therefore, we apply *singular value decomposition* (SVD) directly to the transition matrix instead of the covariance matrix of its elements (as PCA would do), omitting the requirement of centering the values. This way the matrix the SVD is applied to only depends quadratically on the alphabet size $|A|$ and can be stored in a sparse format before the computation. This procedure is also called *latent semantic analysis* [30]. Similar to PCA, we have to pick the dimension of the smaller subspace, which is part of the hyperparameter selection in subsection 5.2.

## 4.5. Classification

The actual classification is done by the $k$-NN algorithm (subsection 3.3), which was selected because of its lazy learning property. Usually ML classifiers operate in two phases, the learning phase and the classification phase. During the learning phase (sometimes also referred to as the training phase), the classifier builds its internal model from the data, which it then uses in the classification phase to classify new data. *Eager learners* (such as neural networks or random forests) have a computationally expensive learning phase, but are very fast during the classification phase. *Lazy learners*, on the other hand, have (almost) no learning phase, but a more computationally expensive classification phase instead. In our case, $k$-NN's lazy learning property is important because of the data's temporary nature. Each time a program receives a software update, the generated events of two identical executions before and after the software update may differ. Therefore, eager learners would have to redo their learning phase to realign their internal model to the program's software update. Since hundreds or even thousands of different programs may be included in a data set, this may cause a drastic performance overhead. Lazy learners, on the contrary, are not affected by this issue since they do not include any kind of preprocessing during the learning phase. This means unlike neural networks and other eager learners, $k$-NN does not need to do additional computation in advance.

The core component of $k$-NN is its distance function. The *Euclidean distance* is most commonly used for that purpose, but since we used `scikit-learn` [31] as our ML framework, the *Minkowski distance*

$$d(x,y) = \left( \sum_{i=1}^{n} |x_i - y_i|^p \right)^{\frac{1}{p}}$$

is the default metric. (Notice that the Minkowski distance is a generalization of the Euclidean distance, which reappears from the equation above by setting $p = 2$.) Since the parameter $p$ is freely choosable, we included it in our hyperparameter search. Because the Minkowski distance weights every element of the feature vector equally, normalization is usually applied to the elements of the feature vector before it is handed over to the classifier. In this case, however, the transition probabilities of the Markov chain are already in the interval $[0, 1]$ and thus no further normalization is required.

## 5. Evaluation

This section describes the data we used for evaluating our system, elaborates on the hyperparameter selection for the algorithms described in section 4 and closes with the evaluation results using four different metrics.

## 5.1. Evaluation Data

The data we used for evaluation of this system was recorded on 19 different hosts over a period of six months and includes a total of 1 138 547 303 events split up into event types as follows:

- 25 886 789 process events
- 256 028 977 registry events
- 218 620 432 image load events
- 499 648 368 file events
- 138 362 737 network events

These events originated from 648 different programs which created 8 797 255 distinct processes. From these 8.7 million executions we excluded those processes which had fewer than six characters after the initial transformation (including time characters), because these did not contain enough relevant information for classification. The total amount of remaining executions stands at 7 858 873. We split this data set into a training set and a verification set at a ratio of 3 to 1, yielding 5 894 376 executions for training and 1 964 497 executions for verification. Because the number of created processes per program depends on multiple factors (such as the program itself, user interaction, network environment, ...), the data set is very imbalanced. Consequently, the split sets are stratified by enforcing the ratio of 3 to 1 for each of the 648 programs.

## 5.2. Hyperparameter Selection

Before conducting the evaluation of our proposed system, we determined the optimal hyperparameters for the algorithms. Hyperparameters are parameters of classifiers, feature extractors, etc. whose choice depends on the data and are usually discovered using a trial-and-error approach. We applied the search in the hyperparameter space using the training set and a threefold stratified cross-validation. Four different hyperparameters were tuned this way: the number of neighbors, the method of majority voting and the distance function (concretely, the $p$ in the formula for the Minkowski distance) for the $k$-NN algorithm, and the number of components for the SVD algorithm. The following possible values were considered for these hyperparameters:

- number of neighbors: $1, 5, 20, 100$
- majority voting method: uniform, distance-weighted
- $p$ in the Minkowski distance: $1, 2, 3$
- number of components: $5, 10, 15, 25, 50, 100$

We used the $F_1$ *score* to evaluate the hyperparameter choices because it is less sensitive to class imbalance than other scores (e. g. accuracy, cf. below). The hyperparameter search resulted in the following values for our parameters:

- number of neighbors: 1
- majority voting method: distance-weighted
- $p$ in the Minkowski distance: 1
- number of components: 100

## 5.3. Results

We used four common data science metrics (namely *accuracy* ($acc$), *precision* ($prec$), *recall* ($rec$) and the $F_1$ score) to determine whether the system is able to correctly classify the system events. These measures are built upon the four basic values of a binary classification task: true positives ($TP$), false positives ($FP$), true negatives ($TN$) and false negatives ($FN$).

$$acc = \frac{TP+TN}{TP+FP+TN+FN} \quad prec = \frac{TP}{TP+FP}$$
$$rec = \frac{TP}{TP+FN} \quad F_1 = 2 \cdot \frac{prec \cdot rec}{prec+rec}$$

Since we are dealing with a multiclass classification, there are different possible methods for averaging these four scores:

- *macro-averaging*: the accuracy, precision, recall and $F_1$ scores are calculated separately for each class, then the mean of these separate scores is taken (without taking class imbalance into account);
- *weighted macro-averaging*: similar to macro-averaging, but weighting the separate scores in the mean calculation according to the number of true instances per class;
- *micro-averaging*: the $TP$, $FP$, $FN$ and $TN$ counts are calculated globally and a single accuracy, precision, recall and $F_1$ score is calculated.

The total values for $TP$, $FP$, $FN$ and $TN$ (as used in micro-averaging) are:

- $TP$: 1 936 312
- $FP$: 8 809
- $FN$: 8 809
- $TN$: 1 186 515 001

In total, only 0.45% of all executions were incorrectly classified.

The results according to the four evaluation scores and the three averaging methods are displayed in Table 3.

| averaging | accuracy | precision | recall | $F_1$ |
|---|---|---|---|---|
| macro | 0.9959 | 0.6140 | 0.5778 | 0.5786 |
| weighted macro- | 0.9959 | 0.9959 | 0.9959 | 0.9959 |
| micro | 0.9959 | 0.9959 | 0.9959 | 0.9959 |

Table 3. Four different evaluation metrics were calculated to evaluate the proposed system: accuracy, precision, recall and $F_1$ score. Because the system task is a multiclass classification, different averaging methods can be used. All four metrics were calculated for macro-, weighted macro- and micro-averaging.

In an effort to visualise the $k$-NN classifier's base data, we projected the transformed data onto a two-dimensional subspace. Concretely, we changed the parameter of the dimensionality reduction step in subsection 4.4 from 100 to 3, giving us the three linear combinations of transition probabilities the SVD deemed most significant. We subsequently plotted a subset of the data we used for evaluation for every choice of two out of these three linear combinations, shown in Figure 4, Figure 5 and Figure 6. Figure 5 shows the data with regard to the first and second most significant linear combination, Figure 6 with regard to the second and third most significant linear combination and Figure 4 with regard to the first and third most significant linear combination. Note that the first mentioned linear combination in the previous sentence is the $x$-axis in the corresponding figure and the second one the $y$-axis, which means that Figure 4 and Figure 5 share the same values on the $x$-axis and Figure 4 and Figure 6 share the same values on the $y$-axis. Each point in these figures represents one process and each different color a different program.

## 6. Discussion

The results in Table 3 show that it is possible to classify programs based on their generated events. Furthermore, the transition probabilities between these events are a viable option as features for such a classification task. Because the evaluation data contains 648 distinct programs, there are also 648 possible classes a new data point can be assigned to. Since this is quite a large number of classes for a multiclass classification task, the worse results using macro-averaging are not too surprising, especially considering the significant class imbalance.

Moreover, the figures show that even in a two-dimensional space the data tends to form homogeneous clusters, making it easier to classify new data points correctly. Based on the results in Table 3, we assume that this effect is even more noticeable in a 100-dimensional space.

The system presented in this paper can already detect anomalies under certain conditions – namely, whenever an intrusion or compromise alters the events a program generates in such a way that the list of generated events after the alteration is more similar to a different program than the one it actually originated from. The detection process is as follows: Naturally, every process has an attached name (the name of the executable file the process spawned from), and every process generates events during its lifetime when it interacts with its software environment. The system proposed in this paper infers from a list of system events the program which is most likely to have generated the event list. If a mismatch between the program name which spawned the process (and therefore generated the event list) and the program name assigned by our proposed system occurs, then an anomaly involving this process has been detected. The system has its limitations, however; while this detection can strongly suggest possible malicious actions, it is not necessarily a guarantee for a compromise having occurred, as the mismatch might also be the result of local noise. Furthermore, the proposed system also cannot guarantee that no malicious actions involving a certain process have occurred even if the assigned program name and the natural program name match.

### 6.1. Limitations

Like any classification system, our proposed system is limited by the underlying data. Since the goal of this paper was to discuss whether a list of events can be assigned to one program with high confidence, the underlying data is precise enough to elaborate on this question. However, an anomaly detection system solely built on this approach might struggle to correctly detect anomalies whenever the anomalous action is only visible at the level of system calls and not at the level of system events. Not every system call leads to an event; normally, there are significantly more system calls than system events and therefore valuable attack information might be lost by only processing certain classes of system events.

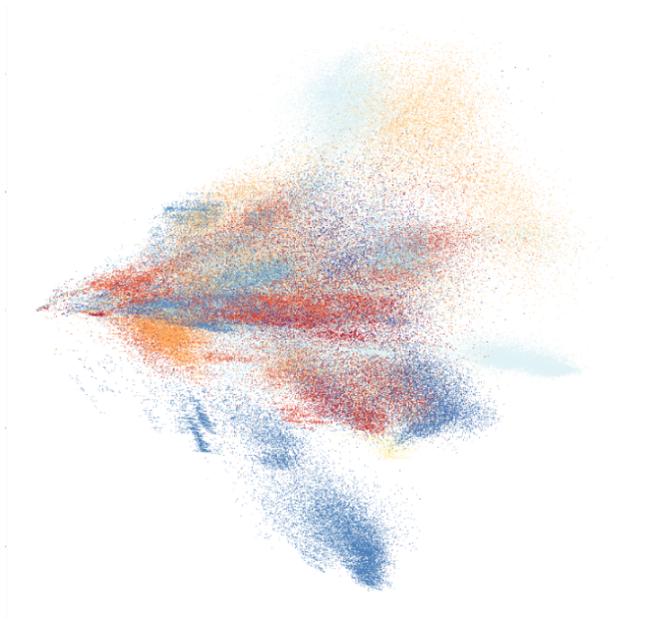

Figure 4. Plot of a subset of the evaluation data according to the first (*x*-axis) and third (*y*-axis) most significant linear combinations of transition probabilities. Each point represents one process and each color a different program.

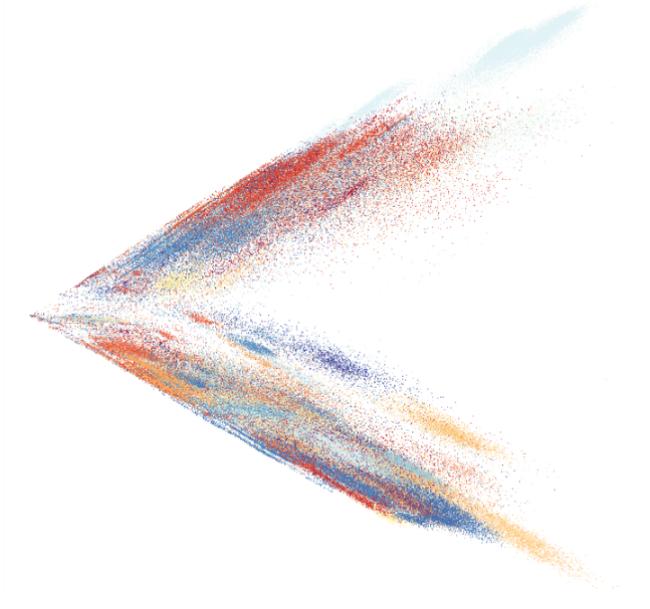

Figure 5. Plot of a subset of the evaluation data according to the first (*x*-axis) and second (*y*-axis) most significant linear combinations of transition probabilities. Each point represents one process and each color a different program.

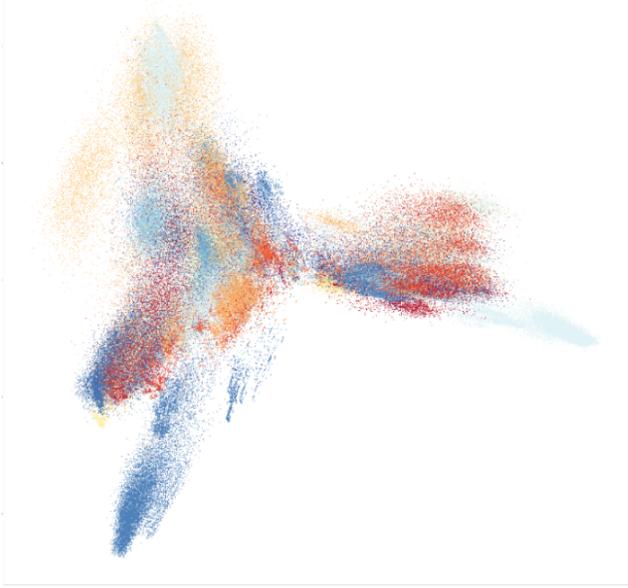

Figure 6. Plot of a subset of the evaluation data according to the second (*x*-axis) and third (*y*-axis) most significant linear combinations of transition probabilities. Each point represents one process and each color a different program.

The data recorded for evaluating this system was collected on several machines of a company with a strong IT development background. Therefore, the way people interacted with programs and the events generated by these programs might differ from common events in a more general setting. Although we believe the generalizations performed in this paper (using event traces, grouping equivalent events, considering multiple neighbors for the majority voting) are sufficient for classifying new data, it is still possible for the model to overfit – while it does not seem to overfit to the data (since it performed quite well on unseen verifica-tion data), it might overfit to the background the training and verification data sets were taken from. Additional data from a different background would be necessary to further improve the confidence in the results.

## 6.2. Future Research

The model proposed in this paper only considers complete executions of processes, consisting of all events from the start of a process until its termination. We would like to investigate further whether it is necessary to restrict ourselves to complete executions or whether including all events within a certain time span irrespective of completion status is also feasible.

Additionally, we still see room for improvement in the alphabet $A$. We deliberately chose the alphabet more general to prevent overfitting. It seems possible that a more fine-grained alphabet choice could achieve even better results while still avoiding overfitting, but additional data from more varied backgrounds would be necessary to provide an accurate answer to this question.

## 7. Conclusion

This paper shows that it is possible to use a list of system events to infer the program which generated these events during its execution. Our proposed system classifies the system event lists using transition probabilities of non-equivalent events as features and by applying a lazy learning algorithm. This classification system can be applied to find mismatches between the classified program name of a list of system events and the name of the program which actually generated these events, thereby detecting anomalies. Hence the system can be used to filter lists of system events as a first stage for a process-based anomaly detection system.

## Acknowledgements

The financial support by the Christian Doppler Research Association, the Austrian Federal Ministry for Digital and Economic Affairs and the National Foundation for Research, Technology and Development is gratefully acknowledged.

The work presented in this paper was done at the Josef Ressel Center for Unified Threat Intelligence on Targeted Attacks (TARGET). TARGET is operated by the St. Pölten University of Applied Sciences.